%% file: root.tex
\documentclass[a4paper, 10pt, conference]{ieeeconf}      % Use this line for a4 paper

\IEEEoverridecommandlockouts                              % This command is only needed if 
\usepackage{amsmath}    % used for equation*
\usepackage{multicol}    % used for equation*
\usepackage{todonotes}  % used for todo notes
\usepackage{multirow}   % used for multirows in
\usepackage{subcaption} % used for subfigures
\usepackage{siunitx}    % used for SI units
\usepackage{array}      % used for constant width in tables
\usepackage{tabu}       % used for force table width to column
    \newcolumntype{C}[1]{>{\centering\arraybackslash}p{#1}}
    \newcolumntype{P}[1]{>{\centering\tiny\arraybackslash}p{#1}}
\usepackage{layouts}
\usepackage[shortcuts]{extdash} % used to avoid linebreaks

\newcommand\copyrighttext{%
	\footnotesize \copyright{ }2020 IEEE. Personal use of this material is permitted. Permission from IEEE must be obtained for all other uses, in any current or future media, including reprinting/republishing this material for advertising or promotional purposes, creating new collective works, for resale or redistribution to servers or lists, or reuse of any copyrighted component of this work in other works.}
\newcommand\copyrightnotice{%
	\begin{tikzpicture}[remember picture,overlay]
	\node[anchor=south,yshift=10pt,xshift=7pt] at (current page.south) {\parbox{\dimexpr\textwidth-\fboxsep-\fboxrule\relax}{\copyrighttext}};
	\end{tikzpicture}%
}
\title{\LARGE \bf
High-Precision Digital Traffic Recording with Multi-LiDAR Infrastructure Sensor Setups*
}

\author{Laurent Kloeker$^{1}$, Christian Geller$^{1}$, Amarin Kloeker$^{1}$ and Lutz Eckstein$^{1}$% <-this % stops a space
\thanks{*The research leading to these results is funded by the European Regional Development Fund (ERDF) within the project “HDV-Mess - High-precision digital traffic recording as a basis for future mobility research - Construction of mobile and modular measuring stations”. The authors would like to thank the consortium for the successful cooperation.}% <-this % stops a space
\thanks{$^{1}$The authors are with the research area Vehicle Intelligence \& Automated Driving, Institute for Automotive Engineering, RWTH Aachen University, 52074 Aachen, Germany
        {\tt\small \{laurent.kloeker, christian.geller, amarin.kloeker, lutz.eckstein\}@ika.rwth-aachen.de}}%
}

\begin{document}

\let\oldtwocolumn\twocolumn
\renewcommand\twocolumn[1][]{%
    \oldtwocolumn[{#1}{
    \begin{center}
           \includegraphics[width=\textwidth]{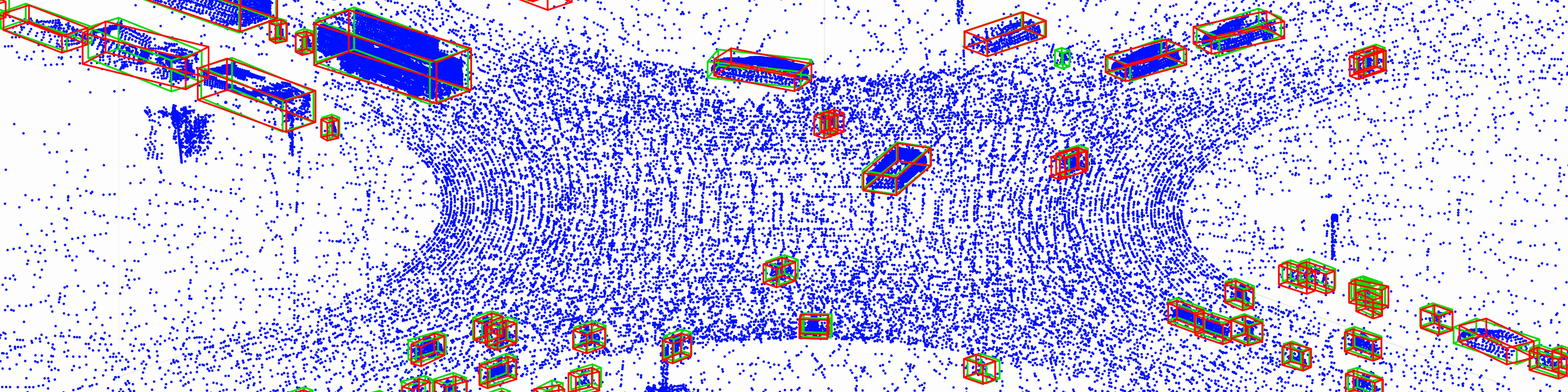}
           \captionof{figure}{Fused point cloud of a multi-LiDAR infrastructure sensor setup at an intersection with tracking results (red) and ground truth data (green), respectively, marked by oriented bounding boxes. The blue point cloud originates from eight LiDAR sensors.}
           \label{fig:bounding_boxes}
        \end{center}
    }]
}

\maketitle
\thispagestyle{empty}
\pagestyle{empty}
% IEEE copyright PART 2 of 2 START
\copyrightnotice
% IEEE copyright PART 2 of 2 END

%%%%%%%%%%%%%%%%%%%%%%%%%%%%%%%%%%%%%%%%%%%%%%%%%%%%%%%%%%%%%%%%%%%%%%%%%%%%%%%%
\begin{abstract}
    Large driving datasets are a key component in the current development and safeguarding of automated driving functions. Various methods can be used to collect such driving data records. In addition to the use of sensor equipped research vehicles or unmanned aerial vehicles (UAVs), the use of infrastructure sensor technology offers another alternative. To minimize object occlusion during data collection, it is crucial to record the traffic situation from several perspectives in parallel. A fusion of all raw sensor data might create better conditions for multi-object detection and tracking (MODT) compared to the use of individual raw sensor data. So far, no sufficient studies have been conducted to sufficiently confirm this approach. In our work we investigate the impact of fused LiDAR point clouds compared to single LiDAR point clouds. We model different urban traffic scenarios with up to eight 64-layer LiDARs in simulation and in reality. We then analyze the properties of the resulting point clouds and perform MODT for all emerging traffic participants. The evaluation of the extracted trajectories shows that a fused infrastructure approach significantly increases the tracking results and reaches accuracies within a few centimeters.
\end{abstract}

%%%%%%%%%%%%%%%%%%%%%%%%%%%%%%%%%%%%%%%%%%%%%%%%%%%%%%%%%%%%%%%%%%%%%%%%%%%%%%%%
\section{INTRODUCTION}
\label{section1}
    
    Large driving datasets form a basis for the development and safeguarding of highly automated driving functions. Only through a huge amount of recorded real traffic data the human driving behavior can be sufficiently represented and used in simulation for further development and validation purposes. The collection of such driving datasets can be implemented in different ways. OEMs, for example, equip their research vehicles with additional sensors that capture the environment from a vehicles perspective \cite{Sun2019, Cordts2016}. Unfortunately, these vehicles are not able to cover their entire surroundings completely. Mounting the sensors on the research vehicles results in a limited field of view (FOV) and can obscure surrounding road users. A second approach to traffic recording is the use of UAVs \cite{Krajewski2018, Bock2019, Zhan2019}. They offer the possibility to record traffic from a bird's eye view, which completely avoids the problem of limited visibility and occlusion. However, the disadvantage of battery-powered UAVs is that the duration of the recordings depends on their battery capacity. As a result, recording times of just \num{20}~minutes are achievable. Infrastructure sensor systems offer a third form of traffic recording, bypassing the disadvantages of the previous two methods. If the sensors are mounted at an appropriate height, such as at traffic lights or lampposts, road users can be recorded with very low or minimal occlusion. As soon as the initial effort of the installation is overcome, the system continuously delivers data of road users and is therefore able to generate large driving datasets.
    
    Past and current infrastructure sensor test fields within the framework of research projects, such as Ko-PER \cite{Meissner2010}, AIM \cite{Aim2020}, test field Lower Saxony \cite{Testfieldlowersaxony2020} and ICT4CART \cite{Ict4cart2020} already apply infrastructure sensor technology for traffic detection. Besides the use of stationary measuring stations, mobile measuring stations can also be used for traffic detection. An advantage of mobile measuring stations is the temporary observation of road cross sections, without having to implement larger construction projects in advance. The results of this work are created within the research project \emph{HDV-Mess}, in which mobile measuring stations are used for high-precision digital recording of road users \cite{Hdvmess2020}.
    \\
    \\
    To minimize object occlusion, it is crucial to record the considered measurement cross section in parallel from several perspectives with the infrastructure sensor system. A highly accurate fusion of the sensor raw data creates a three-dimensional image of the measurement cross section, which has an increased information density compared to the individual sensor raw data. In the past, however, it has not yet been sufficiently proven which advantage a multi-sensor infrastructure setup shows compared to a single-sensor infrastructure setup in the context of high-precision traffic detection.

    In the scope of this work, we focus on the use of LiDARs as infrastructure sensors. We model different traffic scenarios both in simulation and in reality, using between four and eight 64-layer LiDARs. The LiDARs are positioned at different elevated points of the measured cross sections and are directed towards the traffic. In order to generate real test conditions in the simulation data as well, we carry out all experiments not only under optimal sensor conditions but also under nearly realistic sensor conditions. Subsequently, both the individual resulting LiDAR point clouds and the fused point clouds of the entire measurement cross sections are transferred to a MODT algorithm. The extracted trajectories are then compared with a ground truth (GT) and evaluated (see Fig. \ref{fig:bounding_boxes}). The determination of the GT in the simulation data is conducted via the automated output of the actual trajectories and classifications of all road users. In the real measurements a UAV reference system is used, which can extract highly accurate trajectories of all road users of the measurement cross section from an aerial perspective. All extracted trajectories from simulation and real measurements are finally evaluated in order to quantify the advantages of a fused infrastructure sensor setup compared to single sensor setups.

    Our main contributions are as follows:
    \begin{itemize}
        \item We use simulation to build multiple realistic urban traffic scenarios and conduct intensive experiments and comparisons to evaluate the detection accuracy between different sensor setups both under optimal and realistic sensor conditions.
        \item For evaluation, we use a multitude of road user classes: pedestrians, bicycles, motorcycles, cars and trucks.
        \item We reconstruct one of the simulation scenarios on a private test track and compare the the real measurements with those from simulation.
        \item We use a UAV equipped with a high-resolution camera as reference measurement device for the real sensor measurements.
    \end{itemize}

%%%%%%%%%%%%%%%%%%%%%%%%%%%%%%%%%%%%%%%%%%%%%%%%%%%%%%%%%%%%%%%%%%%%%%%%%%%%%%%%
\section{RELATED WORK}
\label{section2}

\subsection{Using LiDAR as Infrastructure Sensors}

    MODT in LiDAR point clouds is a common research topic. Primarily, sensor setups mounted on research vehicles are considered, which record the surrounding traffic, as shown, for example in \cite{Sualeh2019}. A first approach, which used several LiDARs at elevated positions at a traffic intersection as infrastructure sensors, is presented in \cite{Meissner2010}. Up to four 4\=/layer LiDARs were used for detection and tracking of road users. Due to the low sensor resolution the road users could only be roughly tracked. A further approach, which limits itself to the detection and tracking of persons with LiDARs, is shown in \cite{Wenzl2013}. Although LiDARs are not explicitly used as infrastructure sensors in traffic scenarios, the fused sensor information is utilized to get information about the person trajectories. The relatively small measurement cross section of only \(\num{20} \times \num{10}\)~meters poses a limitation for the adaption of this work into a realistic urban traffic scenario. Furthermore, it was only tested with two recorded persons moving through the measurement cross section. Recent work \cite{Herrmann2019} also uses LiDARs as infrastructure sensors at traffic intersections. In addition to real experiments, scenarios are also simulated, as in our approach. Unfortunately, only limited evaluations are made here, since just three vehicles are simulated at a single T-intersection. Different road segments or different classes of road users, such as pedestrians, bicycles and motorcycles, are not considered. Consequently, no comprehensive evaluation is feasible for an adaptation of this approach to real urban traffic scenarios. Additionally to simulation, the real traffic experiments carried out in \cite{Herrmann2019} do not provide any reference data as a basis for evaluation and therefore do not allow any statement about their actual functionality.

    Detached from static infrastructure sensor technology, \cite{Tarko2017} equips a van with a telescopic mast, with a 64-level LiDAR installed at its head for traffic detection. As previously explained, approaches that record a measurement cross section only from one perspective are prone to object occlusion. A general investigation at which locations in urban areas infrastructure sensors have to be positioned in order to achieve an optimal coverage of the considered road segments is presented in \cite{Geissler2019}. Our approach includes modeling these findings on a small scale.

\subsection{Reference Measurement Systems}

    Outside the simulation, a highly accurate and well-proven measurement system is required to generate reference data for real-world experiments. One system that meets those requirements is a camera equipped drone, which measures the multi-LiDAR test scene from a bird's eye view as shown in \cite{Krajewski2019}. The aerial perspective in combination with a high resolution camera allows a consistent detection of all road users to generate highly accurate reference data \cite{Krajewski2018, Bock2019}.

%%%%%%%%%%%%%%%%%%%%%%%%%%%%%%%%%%%%%%%%%%%%%%%%%%%%%%%%%%%%%%%%%%%%%%%%%%%%%%%%
\section{METHOD}
\label{section3}

\subsection{Toolchain}\label{sub:toolchain}
    
    \begin{figure}[tb]
        \centering
        \input{figs/flow_chart.tex}
        \caption{Toolchain overview for our MODT approach.}
        \label{fig:toolchain_overview}
    \end{figure}
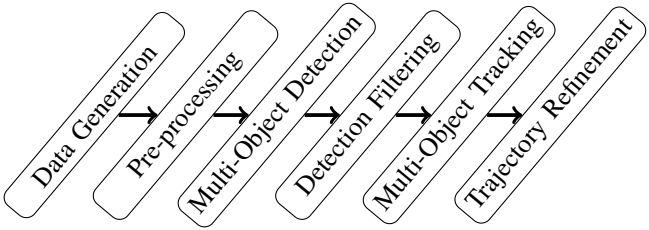

     The following section schematically describes our toolchain for detecting and tracking road users in point clouds (see Fig.~\ref{fig:toolchain_overview}). After generation in simulation or in real measurements, the point clouds are pre-processed. The main goal of this pre-processing step is to minimize the number of points without losing information about the representation of the objects. Therefore, the point cloud is restricted to the region of interest, i.e. the relevant traffic intersection, and irrelevant points on the ground are downsampled. To further reduce the amount of points, zero intensity points are removed before normalizing the total intensity of the dataset.

    For object detection, we use a state-of-the-art deep learning network architecture for 3D point cloud segmentation, named \emph{PointPillars}~\cite{Lang2018}. \emph{PointPillars} is an end-to-end learning approach that divides the point cloud into a pillar grid based on the x-y plane. For each pillar, \(64\) learnable features are extracted, thereby reducing the problem to the 2D plane. This allows the application of a 2D convolutional neural network before a single shot detector generates bounding boxes as network output. All appearing classes of road users are to be detected. Following the classical approach \cite{Ku2018, Zhou2017, Yan2018}, we use two different networks for the detection of vehicles, such as cars and trucks (\texttt{ct}), and vulnerable road users (VRUs), namely pedestrians, bicycles, and motorcycles (\texttt{pbm}). The outputs of \emph{PointPillars} are framewise detections in the shape of oriented 3D bounding boxes. 
    
    In the next step, only detections in the region of interest are considered and duplicated detections are deleted based on their corresponding score. Additionally, consistent trajectories over time are generated on this subset of detections using the tracking algorithm \emph{AB3DMOT}~\cite{Weng2019}. The tracking is based on a Kalman filter with a linear motion model and the Hungarian algorithm for solving the matching problem. A birth and death memory helps to avoid false negatives and false positives and tracks the object consistently over time. 
    The extracted trajectories are refined in a post-processing step in order to generate more smooth and realistic values and filter false detections. A bayesian fixed-interval smoother is applied to all measured variables to smooth the trajectories and therefore significantly reduce the error.

    For the dimensions of the bounding box, we determine fixed sizes for each object based on the corresponding measurement time stamp with the highest point coverage, which typically lies in the scenario center. The heading angle is smoothed by a triangular kernel window, which results in better trajectories and improves accuracy. 

\subsection{Simulation Measurements}
    To generate high-density LiDAR point clouds, we use a given software for driving simulations named \emph{Virtual Test Drive (VTD)}~\cite{Neumann-Cosel2009}. The software allows generic modeling of the environment in both urban and rural scenarios. With an additional ray-tracing plugin, we integrate a model of our used LiDARs, \emph{Ouster OS1} with \(64\) layers, into the VTD environment to record point clouds in different scenarios. We model four scenarios, wherein each eight sensors are pairwise attached at four selected locations at the height of six~meters, one tilted by \num{0.1}~radian and the other by \num{0.3}~radian. Additionally, the sensor orientation is aligned with the scenario center. 
    
\begin{figure}[tb]
        \centering
        \begin{subfigure}{0.48\columnwidth}
            \includegraphics[width=\columnwidth]{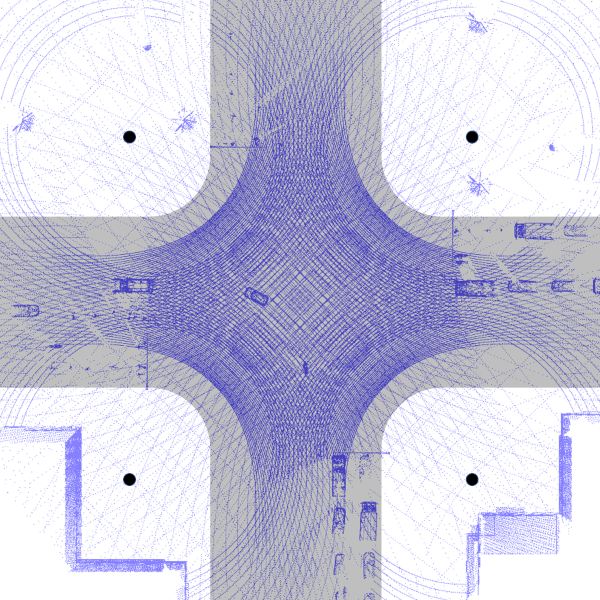}
            \caption{intersection}
        \end{subfigure} 
        \hfill
        \begin{subfigure}{0.48\columnwidth}
            \includegraphics[width=\columnwidth]{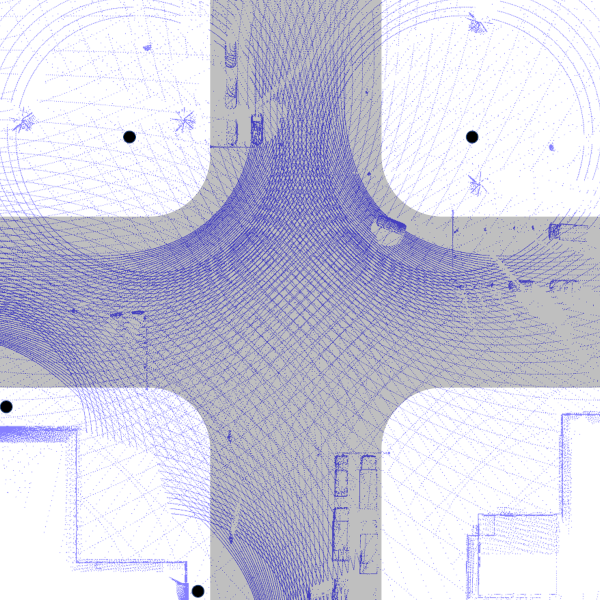}
            \caption{asym. intersection}
        \end{subfigure} 
        \begin{subfigure}{0.48\columnwidth}
            \includegraphics[width=\columnwidth]{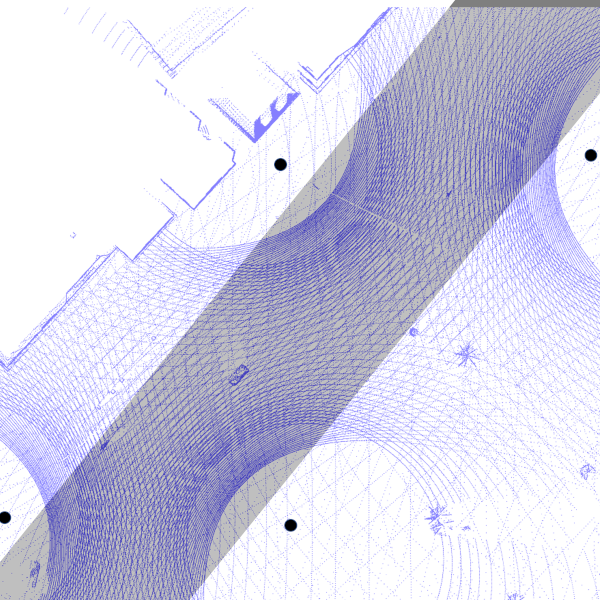}
            \caption{straight}
        \end{subfigure}
        \hfill
        \begin{subfigure}{0.48\columnwidth}
            \includegraphics[width=\columnwidth]{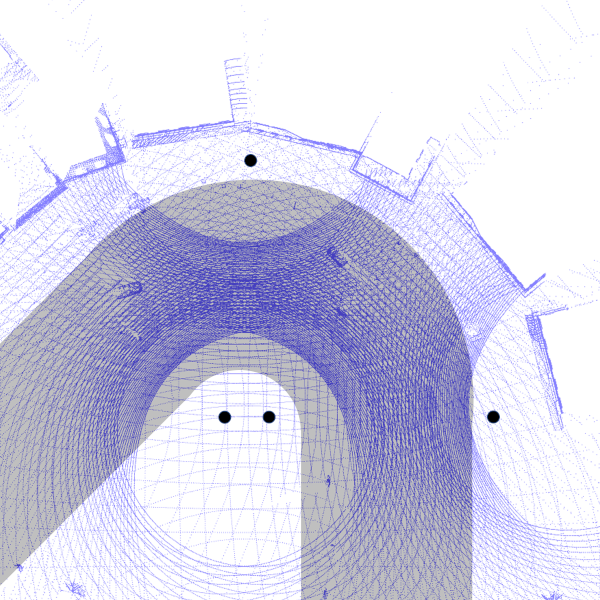}
            \caption{curve}
        \end{subfigure}
        \caption{Fused point clouds of simulated scenarios. Point clouds are generated from all eight sensors. The black circles mark the positions and the gray area the road, respectively.}
        \label{fig:setups}
    \end{figure}

    In the first scenario (A), a symmetric three-lane X-intersection is considered in an urban scenery, with the intersection corners equipped with two LiDAR sensors each. In the second scenario (B), the same setup is used, with one sensor position being changed to avoid the visibility between two of the sensors. Scenario (C) and (D) are constituted of a straight road and a curve with six lanes and one-sided buildings respectively, where the sensors are positioned in a zigzag pattern orthogonal to the road. With the knowledge of the exact sensor pose, the resulting fused point clouds of the described scenarios are shown in Fig.~\ref{fig:setups}.

    In the virtual environment, \(100\) road users, divided into the five classes car, truck, pedestrian, bicycle, and motorcycle, are considered respectively. With a random initial distribution and independent movement, they simulate traffic as realistic as possible. In addition to recording all sensor point clouds, VTD enables the export of GT bounding boxes for all dynamic objects.

    Since the following sections mainly focus on the difference between single and fused point clouds, two separate datasets with point clouds from all of the four scenarios mentioned above are created. The base dataset \texttt{b-s} describes a set consisting of the single point clouds of all sensors, whereas the dataset \texttt{b-f} contains the respective fused point clouds.

    Although the simulation scenarios are modeled in a realistic way, both the ray-tracing mentioned above and the fusion do not include noise as it would appear in the real world. In order to obtain more realistic datasets, a Gaussian distribution of \(\mathcal{N}\left(\SI{0}{\m}, \SI{0.01}{\m^2}\right)\) is applied to each point in an individual point cloud. These values correspond approximately to the noise of our used sensors. The sensor position is interfered with \(\mathcal{N}\left(\SI{0}{\m}, \SI{0.01}{\m^2}\right)\) and the rotational orientation by \(\mathcal{N}\left(\SI{0}{\radian},\SI{2.5e-5}{\radian^2}\right)\) to simulate a noisy point cloud fusion, which results in the equivalent noisy datasets \texttt{n-s} and \texttt{n-f}. All records are cut to a region of interest of \([\SI{-56}{\metre}, \SI{56}{\metre}]\) in x-y-direction and \([\SI{-0.05}{\metre}, \SI{4}{\metre}]\) in z-direction. Furthermore, \num{90}~percent of all points in a five~centimeter (\texttt{b-s, b-f}) or \num{25}~centimeter (\texttt{n-s, n-f}) range around \(z=\SI{0}{\m}\) are removed, since the ground does not provide any added value with respect to object recognition. 
    
    Each of the resulting four datasets (\texttt{b-s, b-f, n-s, n-f}) consists of \(74\,000\) frames, which are divided into \(50\,000\) training and \(24\,000\) evaluation samples recorded with \num{20}~hertz. Since the variety of consecutive samples is comparably low, we reduce the sampling rate of the training set to two~hertz, which results in \(5\,000\) samples. For training with single point clouds, the training dataset consists of point clouds equally distributed from all eight sensors. 

    As described in more detail in the following section, we build up another scenario (E) on our private test track and simulate it in VTD. The training data for this fifth scenario is also provided by VTD. Now, the intersection of scenarios (A) and (B) is equipped with four horizontal sensors at the height of two~meters, and the simulated dynamic objects are reduced to cars, pedestrians, and cyclists only. Similarly, as before, the point clouds contain the same additional noise in point position and sensor fusion, but the ground is completely removed. The relevant area is cut to \([\SI{-40}{\metre}, \SI{40}{\metre}]\) in x-y-direction, due to the modeling of the real test case. As in the previous scenarios, the training data for the test track scenario includes \(5\,000\) point clouds, for each, fused \texttt{t-f} and single point clouds \texttt{t-s}. Tab.~\ref{tab:datasets} illustrates the different datasets in more detail.
    \setlength{\tabcolsep}{6pt}
    \begin{table}[bt]
        \centering
        \caption{Overview of the datasets.}
        \label{tab:datasets}
        \begin{tabu} to \columnwidth{X[c]X[c]X[c]X[c]|X[c]X[c]X[c]X[c]}
        \multicolumn{4}{c|}{\textbf{Scenario A + B + C + D}} & \multicolumn{4}{c}{\textbf{Scenario E}} \\
        \hline
        \multicolumn{4}{c|}{ 8 LiDARs, \(\SI{6}{\m}\) height, tilted} & \multicolumn{4}{c}{4 LiDARs, \(\SI{2}{\m}\) height,  horizontal} \\
        \hline
        \multicolumn{4}{c|}{simulation} & \multicolumn{2}{c}{simulation} & \multicolumn{2}{c}{measurement} \\
        \hline
        \multicolumn{2}{c}{base} & \multicolumn{2}{c|}{noise} & \multicolumn{2}{c}{noise} & \multicolumn{2}{c}{noise} \\
        \texttt{b-s} & \texttt{b-f} & \texttt{n-s} & \texttt{n-f} & \texttt{t-s} & \texttt{t-f} & \texttt{r-s} & \texttt{r-f}
        \end{tabu}
    \end{table}

\subsection{Real Traffic Measurements}
    
\begin{figure}[tb]
        \centering
        \begin{subfigure}{0.49\columnwidth}
            \captionsetup{font=footnotesize, labelfont=footnotesize, labelsep=space, aboveskip=-0.25cm}
            \input{figs/testtrack_overlay.tex}
            \caption{LiDAR setup at test track}
            \label{fig:testtrack_setup}
        \end{subfigure} 
        \hfill
        \begin{subfigure}{0.49\columnwidth}
            \includegraphics[width=\columnwidth]{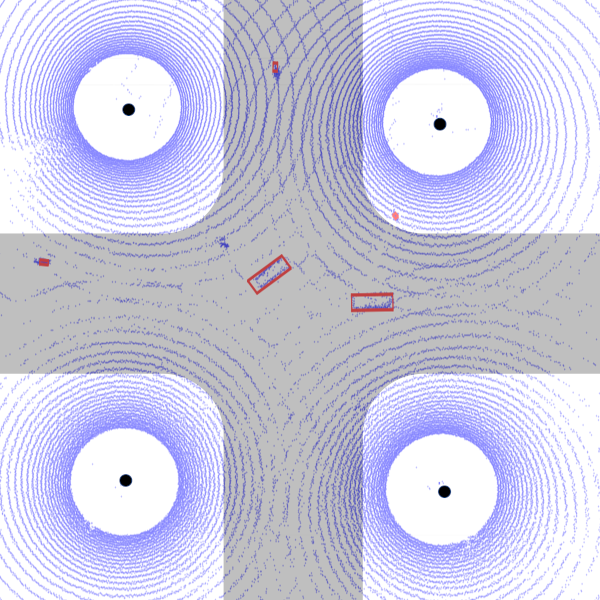}
            \caption{fused measured point cloud}
            \label{fig:testtrack_pc}
        \end{subfigure}
        \caption{Point cloud and test setup of measured scenario. In (a) the sensor position is marked in black, with colored circles roughly symbolizing the FOV of the four LiDARs. The same scenery is shown in the fused point cloud in (b), where bounding boxes highlight the detection results.}
        \label{fig:testtrack}
    \end{figure}

   The previously described scenario (E), is shown from an aerial perspective in Fig.~\ref{fig:testtrack_setup}. We use four \emph{Ouster OS1-64} sensors at \num{20}~hertz, mounted horizontally at the height of two meters and aligned to the intersection center in a distance of \num{28}~meters. Two cars, two pedestrians, and two cyclists move through the crossing area in a period of about \num{12}~minutes. 
    
    We measure the sensor positions and the intersection center point in a local coordinate system and transform them to the UTM-WGS84 coordinate system by matching the points on a geo-referenced orthophoto. Further, the pitch and roll angles are considered to be zero. The yaw angle can be determined from the alignment of the sensors to the intersection center. With this information, a fused point cloud with approximately \(262\,000\) points can be generated. In contrast to the noiseless simulated point clouds, the points are affected by a noise of about \(\mathcal{N}\left(\SI{0}{\m},\SI{0.01}{\m}^2\right)\), which, in addition to measurement inaccuracies of the sensors, can be explained by a noisy fusion, due to deviations in the sensor positions. Analog to the previous procedure, two datasets \texttt{r-f} and \texttt{r-s} with about \(14\,000\) frames each are created for the real test measurement. Fig.~\ref{fig:testtrack_pc} shows the fused, measured point cloud with marked road users in red. 
    
    Equivalently, a pre-processing for the two real datasets \texttt{r-f} and \texttt{r-s} is performed, where the entire ground is removed and the points are cropped to a range of \([\SI{-40}{\metre}, \SI{40}{\metre}]\) in x-y-direction. Further, random noise is filtered out. 

\subsection{Reference Measurements}\label{sub:reference}

    A drone is used to generate highly accurate GT data for the real traffic measurements. The drone videos are recorded using a DJI Phantom 4 Pro. It is equipped with a 4K resolution camera, which has a sampling rate of \num{25}~hertz. Hovering at a height of around \num{100}~meters the whole intersection can be covered. Even though the drone camera is stabilized by a gimbal, drone movements caused by wind are still affecting the recorded data. Therefore, the videos are stabilized in a pre-processing step. Additionally, the distortions caused by the camera lens are corrected in the same step. In order to spatially synchronize the drone data with the LiDAR data, the videos are matched on an orthophoto in the UTM-WGS84 coordinate system using a projective transformation. For the detection of the road users, a neural network performing a semantic segmentation on each video frame is used. The utilized network is derived from a U-Net architecture \cite{Ronneberger2015}. Based on the detections, the trajectory of each road user is extracted and processed, similar to the toolchain described in Section \ref{sub:toolchain}. Using a Kalman filter for the tracking and a bayesian fixed-interval smoother in the post-processing, the centroid, velocity, acceleration and heading of each road user is precisely tracked for each frame. Since the LiDARs' sampling rate is lower than that of the drone camera, the tracks are downsampled to \num{20}~hertz. An initial temporal synchronization between the drone and LiDARs is performed manually by roughly matching the detected bounding boxes. Subsequently, different adjacent frames of the drone data are used to determine the deviation to the LiDAR data. The frame that minimizes the error is used for refining the temporal synchronization.

    The accuracy of the extracted trajectories is determined by the range one pixel covers and by the resolution of the orthophoto the data is matched on. With these values being \(3.5 \times 3.5\) respectively \(10 \times 10\) centimeters, an accuracy of about \num{13.5}~centimeters can be achieved. Taking into account that the stabilization is prone to small errors, the accuracy can worsen up to additional seven centimeters, which equals an error of two pixels. With an optimal temporal synchronization on frame level and a sampling rate of \num{20}~hertz, the maximal time difference \(\Delta t\) between a LiDAR and a drone frame remains \num{25}~milliseconds. This leads to an additional maximal error \(\epsilon_\text{sync}\) of
    \begin{equation}
		\epsilon_\text{sync} = v_\text{max} \cdot \Delta t = \SI{16}{\centi\metre}
    \end{equation}
    at a maximal vehicle speed \(v_\text{max}\) of \num{23}~kilometers~per~hour. The accuracy of the reference data as GT can therefore be specified as \num{13.5} to \num{36.5}~centimeters depending on the velocity of the tracked object and the quality of the stabilization for the considered frame.

\subsection{Evaluation Metrics}

    Our MODT on point clouds is focused on the comparison of single sensor setup and fused sensor setup. Therefore all metrics are applied to both the single and the fused cases in the following.

    Trivially, the density of the points representing an object, and thus the representation of the object, varies with distance to the sensor. In case multiple sensors are used in a setup, not only the amount of points per object generally increases, but also its covered surface. Therefore, we consider the amount of points inside a GT bounding box for each object, which is visualized in a heat map for the respective scenario. Besides the pure amount of points, the dimensions \(\left(h_{m}, w_{m}, l_{m}\right)\) of the minimum bounding box, represented by the occurring points, can be calculated. The ratio to the original dimensions \(\left(h, w, l\right)\) of the GT bounding box then gives information about how well the points represent the box and is visualized in a heat map as well.
    
    Furthermore, as considered in the KITTI Benchmark Suite for Multi-Object Detection and Tracking~\cite{Geiger2012}, we consider the \emph{Average Precision (AP)} as well as quantitative tracking metrics for our detection and tracking pipeline. These are the \emph{Multi-Object Tracking Accuracy (MOTA)} and the \emph{Multi-Object Tracking Precision (MOTP)} introduced by \cite{Bernardin2008}. The MOTA provides information about the accuracy of the predicted bounding boxes against the GT by calculating the averaged distance. The MOTP, on the other hand, indicates whether all trajectories are found, if they are continuous or all trajectories are correctly assigned. 

    As a further metric, the \emph{Mean Averaged Error (MAE)} to the GT in position, velocity, and acceleration is calculated for every dataset. Therefore, we consider a set of chosen trajectories \(T\), which exceed a specific length \(l_t\) and number of frames \(n_t\). The MAE is calculated by the deviations in all corresponding frames of this trajectory set as 
    \begin{equation}
        d_T= \frac{ \sum\limits_{t \in T} \sum\limits_{i=1}^{n_t} d_{t,i}}{\sum\limits_{t \in T} n_t} \quad \textrm{with} \quad  T \in  \{T_{\textrm{veh}}, T_{\textrm{vru}}\} \in \{T_{\textrm{all}}\},
    \end{equation}
    where \(d_{t,i}\) is the deviation of trajectory \(t\) in frame \(i\). The deviation \(d_T\) for the entire trajectory set \({T}_{\textrm{all}}\), as well as the deviation scores for GT vehicles \({T}_{\textrm{veh}}\) and VRUs \({T}_{\textrm{vru}}\) are determined, respectively. Since, only the deviations play a role, misclassifications are neglected.

%%%%%%%%%%%%%%%%%%%%%%%%%%%%%%%%%%%%%%%%%%%%%%%%%%%%%%%%%%%%%%%%%%%%%%%%%%%%%%%%
\section{EXPERIMENTS}
\label{section4}

\subsection{Training of Neural Networks}

    Two neural networks are trained on each of the simulated datasets, one for large objects such as cars and trucks (\texttt{ct}), the other for smaller objects like pedestrians, bicyclists, and motorcycles (\texttt{pbm}). Due to the different applications, the amounts of parameters of the models differ. This results in a reduced batch size of one for the \texttt{pbm} model in comparison to two for \texttt{ct}. All other parameters, such as voxel size, number of points per voxel, learning rate, as well as the architecture with the filter and layer amounts, are adopted from the standard \emph{PointPillars} settings, which we utilize in our approach.
    
    The four simulated scenarios with datasets \texttt{b-s}, \texttt{b-f}, \texttt{n-s}, \texttt{n-f} serve as proof-of-concept, while dataset \texttt{t-s} and \texttt{t-f} are used as training data for the real use case on the test track, due to a comparable setup. The data is divided into \(4\,000\) training and \(1\,000\) validation samples, with each frame containing in average \(25\) objects of all classes. Training is performed on a \emph{NVIDIA Titan RTX} GPU with \num{24}~gigabyte VRAM.
    
    %Tab.~\ref{tab:trained_nets} shows the different networks, the comparable amount of training epochs, and the remaining loss. Although almost the same number of training epochs was trained, the losses behave very differently. In general, fused datasets show smaller losses compared to single datasets, and training on the base case was more effective than training on the noisy datasets.  
    %
    \iffalse
    \setlength{\tabcolsep}{6pt}
    \begin{table}[h]
        \centering
        \caption{Overview over neural networks}
        \label{tab:trained_nets}
        \begin{tabu}to\columnwidth{X[c]|C{0.75cm}C{0.75cm}C{0.75cm}C{0.75cm}C{0.75cm}C{0.75cm}}
        \textbf{Dataset} &\textbf{\texttt{b-s}} & \textbf{\texttt{b-f}} & \textbf{\texttt{n-s}} & \textbf{\texttt{n-f}} & \textbf{\texttt{t-s}} & \textbf{\texttt{t-f}} \\
        \hline
        {Classes} & CT & CT & CT & CT & C & C \\
        {Epochs} & 100 & 117 & 70 & 98 & 82 & 163 \\
        {Loss} & 0.341 & 0.106 & 0.378 & 0.125 & 0.264 & 0.139 \\
        \hline      
        {Classes} & PBM & PBM & PBM & PBM & PB & PB \\
        {Epochs} & 103 & 100 & 70 & 54 & 79 & 126 \\
        {Loss} & 0.248 & 0.098 & 0.305 & 0.223 & 0.289 & 0.163
        \end{tabu}
    \end{table}
    \fi

\subsection{Qualitative and Quantitative Evaluation of Different Sensor Setups}
    
\begin{figure}[tb]
        \begin{subfigure}{0.43\columnwidth}
            \includegraphics[width=\columnwidth, trim={0 0 4.8cm 0}, clip=true]{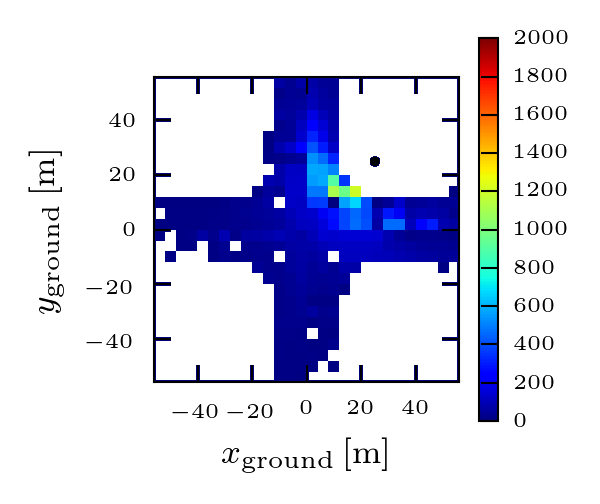}
            \caption{single sensor point cloud}
        \end{subfigure} 
        \hfill
        \begin{subfigure}{0.55\columnwidth}
            \includegraphics[width=\columnwidth]{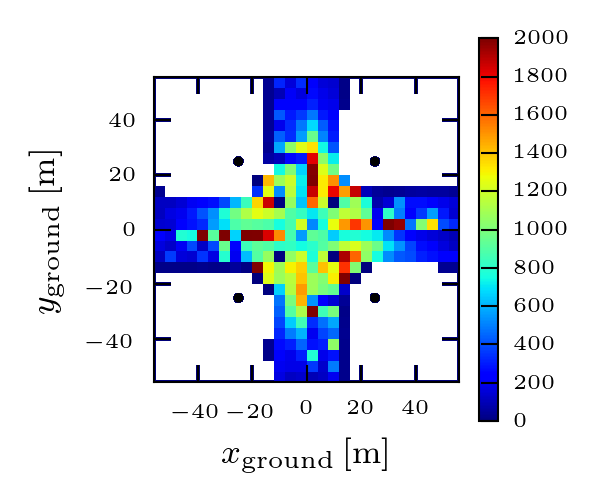}
            \caption{fused sensor point cloud}
        \end{subfigure} 
        \centering
        \caption{Amount of points per bounding box for \(7\,250\) frames of scenario (A). The sensor positions are marked with black circles.}
        \label{fig:heatmap_number_points}
    \end{figure}

    In the following, the difference between the single and multi-sensor setup concerning the point representation of the road users is examined more closely. Scenario~(A) is used as an example, equipped with a single sensor and a multi-sensor scenario with eight sensors respectively. The amount of points of all objects in \(7\,250\) frames is entered in the specific location cell on a heat map with a grid size of four meters in the full range of \([\SI{-56}{\m}, \SI{56}{\m}]\) in x-y-direction and is illustrated in Fig.~\ref{fig:heatmap_number_points}. Due to the arrangement of the single sensor at the upper right intersection corner, higher amounts only occur in this area but are clearly below the achieved amount of points in the fused case, which consists of the eight symmetrically arranged sensors. It is evident that the point density in the merged case is significantly higher, especially in the area of the overlapping intersection.

\begin{figure}[tb]
        \centering
        \begin{subfigure}{0.44\columnwidth}
            \includegraphics[width=\columnwidth, trim={0 0 4cm 0}, clip=true]{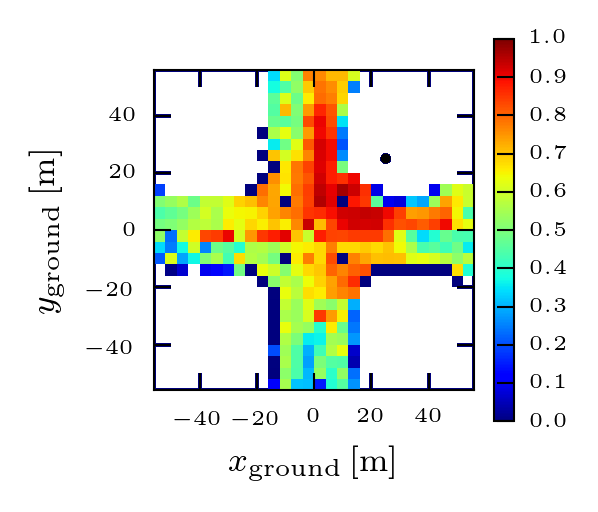}
            \caption{width ratio single}
        \end{subfigure} 
        \hfill
        \begin{subfigure}{0.54\columnwidth}
            \includegraphics[width=\columnwidth]{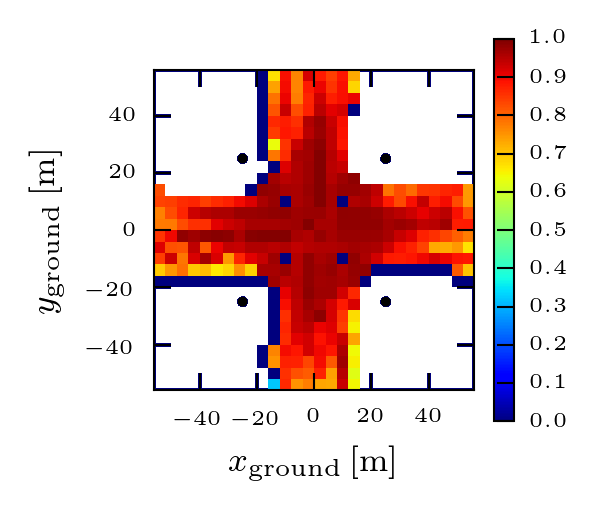}
            \caption{width ratio fused}
        \end{subfigure} 
        \hfill
        \begin{subfigure}{0.44\columnwidth}
            \includegraphics[width=\columnwidth, trim={0 0 4cm 0}, clip=true]{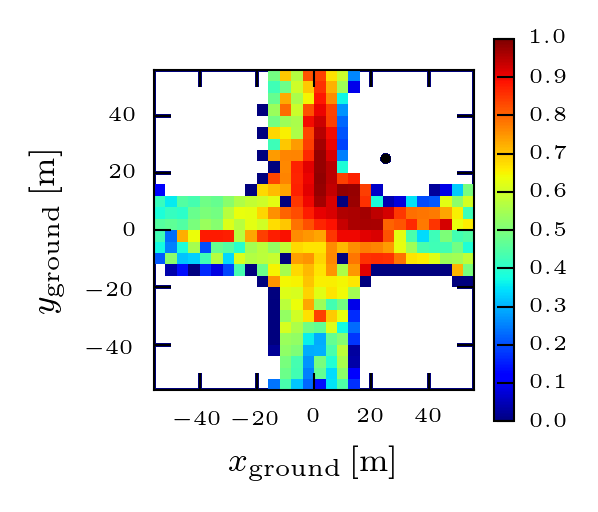}
            \caption{length ratio single}
        \end{subfigure} 
        \begin{subfigure}{0.54\columnwidth}
            \includegraphics[width=\columnwidth]{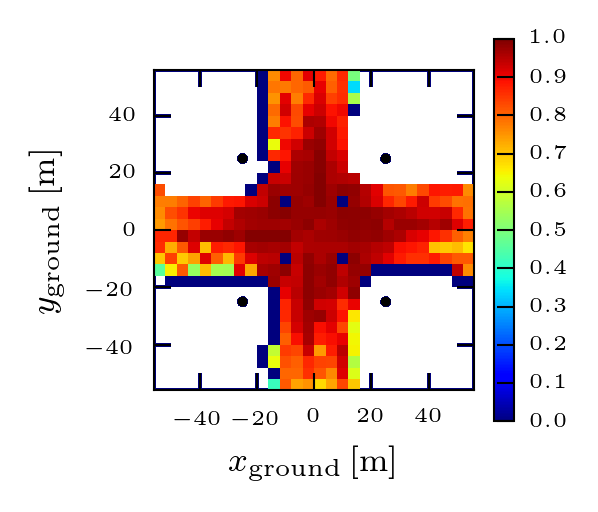}
            \caption{length ratio fused}
        \end{subfigure} 
        \hfill
        \begin{subfigure}{0.44\columnwidth}
            \includegraphics[width=\columnwidth, trim={0 0 4cm 0}, clip=true]{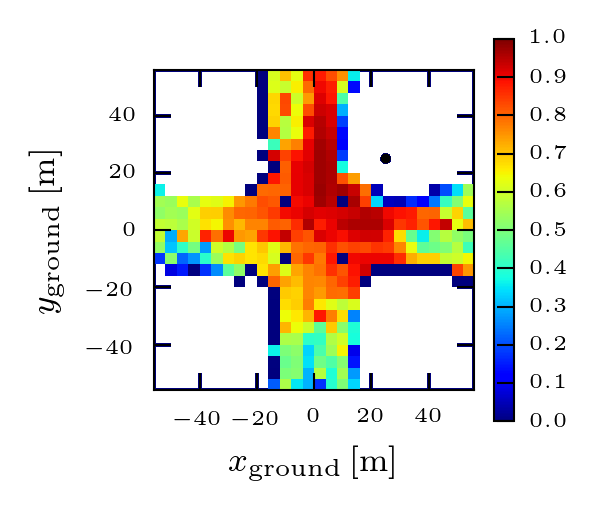}
            \caption{height ratio single}
        \end{subfigure} 
        \hfill
        \begin{subfigure}{0.54\columnwidth}
            \includegraphics[width=\columnwidth]{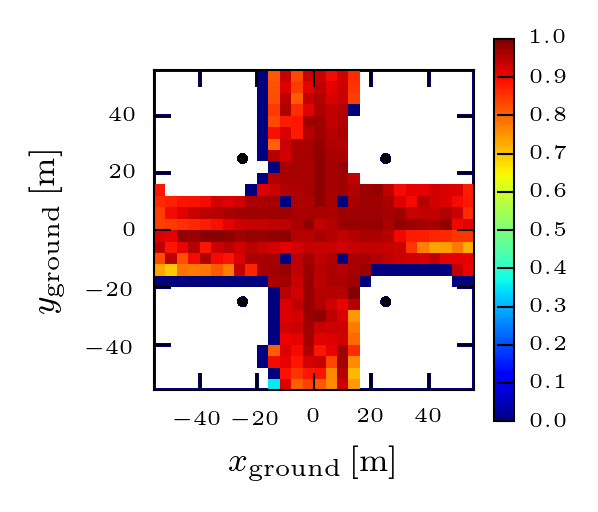}
            \caption{height ratio fused}
        \end{subfigure} 
        \caption{Bounding box coverage in single and fused sensor point cloud for \(7\,250\) frames of scenario (A). The sensor positions are marked with black circles.}
        \label{fig:heatmap_coverage}
    \end{figure}

    Even more interesting is the comparison of a GT bounding box and the minimal bounding box, which is defined by the points describing that object. The ratio in the dimensions of these bounding boxes is examined for \(7\,250\) frames of scenario~(A) and inserted again into a heat map for the single and multi-sensor case, respectively. The results for length, width, and height are shown in Fig.~\ref{fig:heatmap_coverage} for the single sensor and the fused case. Again, it is evident that in a local environment around the single sensor, the relative coverage of the original bounding box is the highest. In the fused case, the area coverage increases due to the higher number of points and the sensor positions. The maximum is reached in the center of the intersection, where all eight sensors are aligned and no occlusion occurs. Furthermore, in both cases, a difference between width, length and height ratio can be detected. Compared to the width and length ratio, the height ratio is relatively well-represented, even at very distant objects, which can be explained by the sensors height of six~meters and the resulting low occlusion. The four less represented cells in the middle of the intersection result from the fact that no road user occurs at this location due to the road structure of the crossing. For reasons of symmetry, the single case is only examined as an example on one sensor.
    \setlength{\tabcolsep}{5pt}
    \begin{table*}[t]       
        \centering 
            \caption{Detection and tracking results of simulation measurements. Base and noisy cases are considered separately, where bold values highlight the highest value of each column.}
            \label{tab:results_sim}
            \begin{tabular}{c|ccc|ccc|ccc|ccc|ccc}
                \multirow{2}{*}{\textbf{Case}} & \multicolumn{3}{c|}{\textbf{Car}} & \multicolumn{3}{c|}{\textbf{Truck}} & \multicolumn{3}{c|}{\textbf{Pedestrian}} & \multicolumn{3}{c|}{\textbf{Bicycle}} & \multicolumn{3}{c}{\textbf{Motorcycle}}  \\
                & \footnotesize AP & \footnotesize MOTA & \footnotesize MOTP & \footnotesize AP & \footnotesize MOTA & \footnotesize MOTP & \footnotesize AP & \footnotesize MOTA & \footnotesize MOTP & \footnotesize AP & \footnotesize MOTA & \footnotesize MOTP & \footnotesize AP & \footnotesize MOTA & \footnotesize MOTP \\
                \hline
                \texttt{b-s} & 0.4197 & 0.6309 & 0.6128 & 0.6536 & 0.6226 & 0.6840 & 0.1932 & 0.2811 & \textbf{0.4543} & 0.2813 & 0.4012 & 0.4853 & 0.5659 & 0.5108 & \textbf{0.4987}  \\
                \texttt{b-f} & \textbf{0.9142} & \textbf{0.8459} & \textbf{0.7249} & \textbf{0.8958} & \textbf{0.7245} & \textbf{0.7721} & \textbf{0.6708} & \textbf{0.6260} & 0.4284 & \textbf{0.7411} & \textbf{0.8050} & \textbf{0.5116} & \textbf{0.8105} & \textbf{0.6831} & 0.4876  \\
                \hline
                \texttt{n-s} & 0.3992 & {0.5467} & 0.6085 & 0.5875 & {0.6031} & {0.6972} & 0.1238 & {0.2259} & 0.4406 & 0.2418 & {0.4234} & 0.4836 & 0.4721 & {0.4754} & {0.5506} \\
                \texttt{n-f} & \textbf{0.8184} & \textbf{0.8497} & \textbf{0.6692} & \textbf{0.9822} & \textbf{0.7495} & \textbf{0.7583} & \textbf{0.5413} & \textbf{0.3458} & \textbf{0.4691} & \textbf{0.6988} & \textbf{0.7202} & \textbf{0.5213} & \textbf{0.8431} & \textbf{0.6470} & \textbf{0.5733}   
            \end{tabular}
    \end{table*}
    
\subsection{Detection and Tracking Results of Simulation Measurements}

    In the following section, the toolchain presented in Section~\ref{section3} is applied to the simulated datasets, and the methods for MODT of road users are analyzed and evaluated. An evaluation of the two respective neural networks is performed, so that tracking is conducted for all classes to build the trajectories subsequently. Each evaluation dataset consists of \(24\,000\) frames based on scenarios (A) - (D). The AP of the detections, as well as MOTA and MOTP after the tracking step, are examined for each class in Tab.~\ref{tab:results_sim}.

    In general, vehicles can be detected more precisely and accurately due to their large amount of representing points. Moreover, it becomes clear that the additional coverage of the objects in the multi-LiDAR setup provides an added value in object detection and tracking. Regardless of the class, the fused case performs better than the single case, but the differences are highest for VRUs and lowest for large objects like trucks. This can be explained by the representation of the objects by the point sets. Even with the single sensor, a truck can be represented well due to its dimensions, which is not possible with a pedestrian, especially at long distances, due to the low amount of detected points.

    Comparing the AP of base and noise datasets, no significant difference can be observed, which means the network achieves good results even with noisy data. However, the variation in the MOTA increases with decreasing object size. The noise has only limited influence on large vehicles but makes tracking more challenging for small road-users such as pedestrians.

    The trajectories smoothed in post-processing are compared with the GT data from the simulation, and a deviation in position, velocity, and acceleration is calculated. The average error of all trajectories with length larger ten~meters and frame amount higher than \num{50} is visualized for all four simulated datasets in Tab.~\ref{tab:differences}. Again, the fused cases show a much higher precision, both in base and noisy datasets.
    %
    % TODO: describe difference between position, acceleration, velocity ?
    % TODO: describe difference between AP MOTA MOTP ?
    %
    \begin{figure}[tb]
        \centering
        \begin{subfigure}{0.49\columnwidth}
            \includegraphics[width=\columnwidth, trim={3.5cm 1.75cm 3cm 1.75cm}, clip=true]{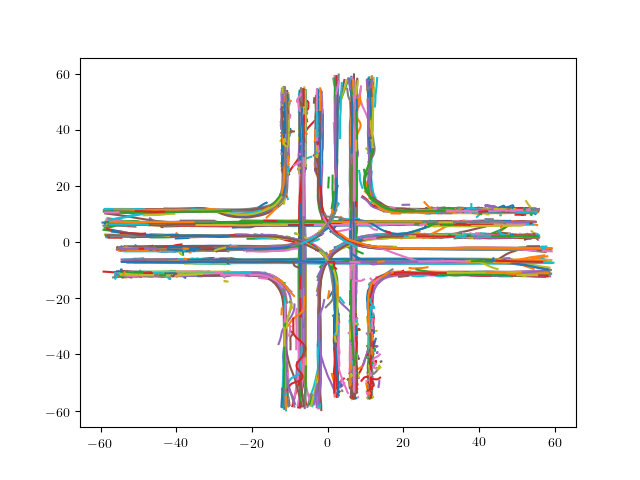}
            \caption{single sensor point cloud}
        \end{subfigure} 
        \begin{subfigure}{0.49\columnwidth}\centering
            \includegraphics[width=\columnwidth, trim={3.5cm 1.75cm 3cm 1.75cm}, clip=true]{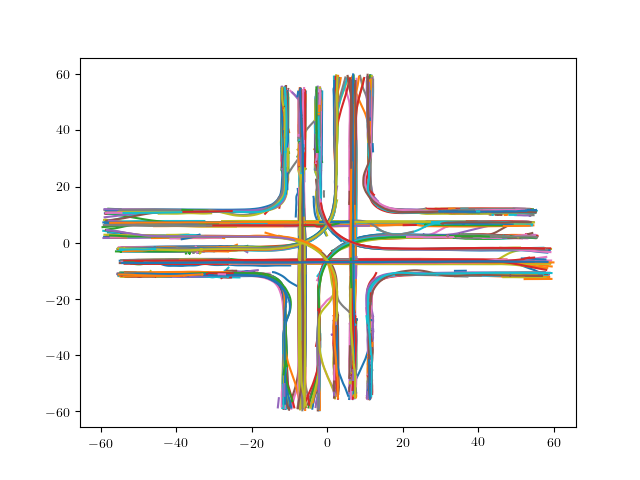}
            \caption{fused sensor point cloud}
        \end{subfigure}
        \caption{Resulting smoothed trajectories.} %Maybe mark starting and end point of each trajectory
        \label{fig:tracking}
    \end{figure}
    Fig.~\ref{fig:tracking} shows the trajectories for an evaluation of scenario~(A) with \(6\,000\) base point clouds for the single and fused case respectively. Although most of the objects are detected in both cases, the precision is more accurate is the fused case. Furthermore, the trajectories in the single case show short lengths, due to missing continuity.

\subsection{Detection and Tracking Results of Real Traffic Measurements}

    Similar to the previous section, we perform detection and tracking on the real, measured point clouds and compare them with the drone data, where the drone trajectories serve as GT. Both measured datasets \texttt{r-s} and \texttt{r-f} are evaluated with the networks trained on the corresponding simulated datasets. Subsequently, tracking over all detections is performed, which results in the metrics in Tab.~\ref{tab:results_real}. 
    \setlength{\tabcolsep}{5pt}
    \begin{table}[bt]     
        \centering
        \caption{Tracking results of real measurements, where bold values highlight the highest value of each column.}
        \label{tab:results_real}
        \begin{tabu} to \columnwidth{X[c]|cc|cc|cc}
            \multirow{2}{*}{\textbf{Case}} & \multicolumn{2}{c|}{\textbf{Car}} & \multicolumn{2}{c|}{\textbf{Pedestrian}} & \multicolumn{2}{c}{\textbf{Bicycle}} \\
            & \footnotesize MOTA & \footnotesize MOTP & \footnotesize MOTA & \footnotesize MOTP & \footnotesize MOTA & \footnotesize MOTP \\
            \hline
            \texttt{r-s} & 0.2202 & \textbf{0.3633} & 0.0071 & 0.3191 & 0.0577 & 0.3034  \\
            \texttt{r-f} & \textbf{0.3056} & 0.3624 & \textbf{0.6692} & \textbf{0.3744} & \textbf{0.2053} & \textbf{0.3260}
        \end{tabu}
    \end{table}
    Similar phenomena as before become visible. With a multi-sensor setup, better detection and tracking results can also be achieved for real applications. Nevertheless, the average precision and the tracking metrics are significantly lower than in the previously investigated base cases, due to training on simulated data and real noisy sensor settings. As before, Tab.~\ref{tab:differences} shows the final average deviations in the real case for chosen trajectories larger than ten meters and with more than \num{50} frames. Similar to the simulation datasets, the deviation to the extracted drone data is much lower when using the multi-sensor setup. If vehicles and VRUs are examined separately, it becomes apparent that the deviations for VRUs are smaller. This can be observed in all datasets and results from the overall smaller dimensions of the road users. With our approach, we achieve an accuracy of \num{0.47}~meters for the multi-LiDAR setup. Considering the deviations of the drone data (see Section \ref{sub:reference}), the results are comparable with those of the simulation. Therefore, our approach achieves a good accuracy with respect to the reality.
    \setlength{\tabcolsep}{6pt}
    \begin{table}[tb]
        \centering       
            \caption{Averaged deviations to GT data in position, velocity and acceleration, averaged over all evaluation datasets, comparing the single and fused case. All trajectories over ten~meters length and \num{50}~frames are considered for the evaluation. Bold entries show the best result for base, noisy, and real data respectively.}
            \label{tab:differences}
            \begin{tabu}to \columnwidth{X[c]X[c]|X[c]X[c]|X[c]X[c]|X[c]X[c]}
                \multicolumn{2}{c|}{\textbf{Deviation}} & \texttt{b-s} & \texttt{b-f} & \texttt{n-s} & \texttt{n-f} & \texttt{r-s} & \texttt{r-f} \\
                \hline
                \multirow{3}{*}{\footnotesize \textbf{pos} \([\SI{}{\m}]\)} & all & 0.38 & \textbf{0.18} & 0.39 & \textbf{0.20} & 0.77 & \textbf{0.47}\\
                 & vehicle & 0.46 & \textbf{0.17} & 0.52 & \textbf{0.20} & 0.97 & \textbf{0.65}\\
                 & vru & 0.30 & \textbf{0.15} & 0.29 & \textbf{0.20} & 0.8 & \textbf{0.29}\\
                \hline
                \multirow{3}{*}{\footnotesize \textbf{vel} \([\SI{}{\frac{\m}{\s}}]\)} & all & 0.33 & \textbf{0.30} & 0.28 & \textbf{0.19} & 0.32 & \textbf{0.24}\\
                & vehicle & 0.41 & \textbf{0.23} & 0.41 & \textbf{0.25} & 0.57 & \textbf{0.5}\\
                & vru & 0.26 & \textbf{0.19} & 0.18 & \textbf{0.17} & 0.28 & \textbf{0.12}\\
                \hline
                \multirow{3}{*}{\footnotesize \textbf{acc} \([\SI{}{\frac{\m}{\s^2}}]\)} & all & 0.72 & \textbf{0.69} & 0.76 & \textbf{0.68} & 0.53 & \textbf{0.45}\\
                & vehicle & 0.81 & \textbf{0.50} & 0.94 & \textbf{0.76} & 1.09 & \textbf{0.99}\\
                & vru & 0.64 & \textbf{0.45} & 0.61 & \textbf{0.66} & 0.44 & \textbf{0.23}
            \end{tabu}
    \end{table}
    
\addtolength{\textheight}{-11.25cm}   % This command serves to balance the column lengths
    
%%%%%%%%%%%%%%%%%%%%%%%%%%%%%%%%%%%%%%%%%%%%%%%%%%%%%%%%%%%%%%%%%%%%%%%%%%%%%%%%
\section{CONCLUSIONS}
\label{section5}

The fusion of LiDAR point clouds enable the use of multiple LiDAR sensor setups for the recording of traffic data. As shown in the evaluation part of this paper, these fused point clouds offer optimized conditions for road user detection and tracking compared to the single sensor case. This is due to a higher relative coverage of road users and takes effect in both, simulation and real measurements. In future work we want to evaluate the same approach for other sensor types such as cameras and radars as well as for a combination of all these sensor types.

%%%%%%%%%%%%%%%%%%%%%%%%%%%%%%%%%%%%%%%%%%%%%%%%%%%%%%%%%%%%%%%%%%%%%%%%%%%%%%%%
% General TODOs:
%   mention no real time processing ?
%   mention that positions in real application are assumed to be constant ? 
%   Fehler von Fusion nennen
%   improvement of fused setup in percentage !

%   Schusterjungen und Hurenkinder

%%%%%%%%%%%%%%%%%%%%%%%%%%%%%%%%%%%%%%%%%%%%%%%%%%%%%%%%%%%%%%%%%%%%%%%%%%%%%%%%

%%%%%%%%%%%%%%%%%%%%%%%%%%%%%%%%%%%%%%%%%%%%%%%%%%%%%%%%%%%%%%%%%%%%%%%%%%%%%%%%

%%%%%%%%%%%%%%%%%%%%%%%%%%%%%%%%%%%%%%%%%%%%%%%%%%%%%%%%%%%%%%%%%%%%%%%%%%%%%%%%
%\section*{APPENDIX}


%\section*{ACKNOWLEDGMENT}

%The preferred spelling of the word ÒacknowledgmentÓ in America is without an ÒeÓ after the ÒgÓ. Avoid the stilted expression, ÒOne of us (R. B. G.) thanks . . .Ó  Instead, try ÒR. B. G. thanksÓ. Put sponsor acknowledgments in the unnumbered footnote on the first page.

%%%%%%%%%%%%%%%%%%%%%%%%%%%%%%%%%%%%%%%%%%%%%%%%%%%%%%%%%%%%%%%%%%%%%%%%%%%%%%%%

\bibliographystyle{IEEEtran} % use IEEEtran.bst style
\bibliography{literature}

\end{document}

%% file: figs/flow_chart.tex
\usetikzlibrary{calc}
\begin{tikzpicture}
    \tikzstyle{flow} = [draw, rectangle, rounded corners, minimum width=3.3cm]
   
    \begin{scope}[shift={(-3,0)}]
        \node [flow, rotate=50] (p1) {Data Generation};
    \end{scope}
    \begin{scope}[shift={(-1.8,0)}]
         \node [flow, rotate=50] (p2) {Pre-processing};
    \end{scope}
        \begin{scope}[shift={(-0.6,0)}]
         \node [flow, rotate=50] (p3) {Multi-Object Detection};
    \end{scope}
    \begin{scope}[shift={(0.6,0)}]
         \node [flow, rotate=50] (p4) {Detection Filtering};
    \end{scope}
    \begin{scope}[shift={(1.8,0)}]
         \node [flow, rotate=50] (p5) {Multi-Object Tracking};
    \end{scope}
    \begin{scope}[shift={(3,0)}]
         \node [flow, rotate=50] (p6) {Trajectory Refinement};
    \end{scope}
    
    \draw[->, line width=0.5mm] (p1) -- (p2);
    \draw[->, line width=0.5mm] (p2) -- (p3);
    \draw[->, line width=0.5mm] (p3) -- (p4);
    \draw[->, line width=0.5mm] (p4) -- (p5);
    \draw[->, line width=0.5mm] (p5) -- (p6);
\end{tikzpicture}

%% file: figs/testtrack_overlay.tex
    \usetikzlibrary{fadings}
    \begin{tikzpicture}
        \tikzstyle{sensor} = [circle, fill=black, minimum size=0.075cm, inner sep=0pt]
        \tikzfading[name=myfading, outer color=transparent!80, inner color=transparent!0]
    
        \node [anchor=south west, inner sep=0] (image) at (0,0) {\includegraphics[width=\columnwidth, angle=180]{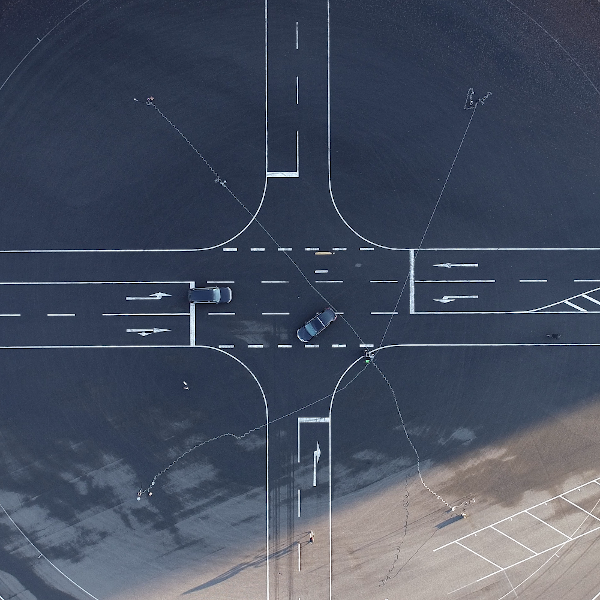}};
        
        \begin{scope}[x={(image.south east)},y={(image.north west)}]
            % clip circles
            \clip (0,0) rectangle (1,1);
            
            \fill[path fading=myfading, cyan, opacity=0.6] (0.23,0.85) circle(2cm);
            \fill[path fading=myfading, green, opacity=0.6] (0.76,0.82) circle(2cm);
            \fill[path fading=myfading, yellow, opacity=0.6] (0.21,0.17) circle(2cm);
            \fill[path fading=myfading, red, opacity=0.6] (0.76,0.17) circle(2cm);
            
            % sensors
            \node [draw, sensor] at (0.23,0.85) {}; % top left
            \node [draw, sensor] at (0.76,0.82) {}; % top right
            \node [draw, sensor] at (0.21,0.17) {}; % bottom left
            \node [draw, sensor] at (0.76,0.17) {}; % bottom right
        \end{scope}
    \end{tikzpicture}